\pdfoutput=1

\documentclass[11pt]{article}

\usepackage{naacl2021}

\usepackage{times}
\usepackage{latexsym}

\usepackage{hyperref}
\usepackage{url}
\usepackage{enumitem}

\usepackage{times}  
\usepackage{helvet} 
\usepackage{courier}  
\usepackage{natbib}  
\usepackage{caption} 
\usepackage{enumitem}
\usepackage{multirow}
\usepackage{booktabs}
\usepackage{subcaption}
\usepackage{xr}
\usepackage{bbold}
\usepackage{booktabs}
\usepackage{amsmath}
\usepackage{xcolor,pifont}
\usepackage{MnSymbol,wasysym}
\usepackage{fontawesome}
\usepackage{graphicx}
\usepackage{anyfontsize}
\usepackage{tikz}
\usepackage{todonotes}
\usepackage{array}

\usepackage[T1]{fontenc}

\usepackage[utf8]{inputenc}

\usepackage{microtype}

\usepackage{amsmath,amsfonts,bm}

\def\Figref#1{Figure~\ref{#1}}

\def\secref#1{section~\ref{#1}}
\def\Secref#1{Section~\ref{#1}}

\def\eqref#1{equation~\ref{#1}}

\def\1{\bm{1}}

\def\mL{{\bm{L}}}

\DeclareMathAlphabet{\mathsfit}{\encodingdefault}{\sfdefault}{m}{sl}
\SetMathAlphabet{\mathsfit}{bold}{\encodingdefault}{\sfdefault}{bx}{n}

\title{\lampret: Layout-Aware Multimodal PreTraining for Document Understanding}

{\centering
    \author{
        Te-Lin Wu$^{*1}$,
        Cheng Li$^2$,
        Mingyang Zhang$^2$,
        Tao Chen$^2$,
        \\
        \textbf{
        Spurthi Amba Hombaiah$^2$,
        Michael Bendersky$^2$
        }
        \\

    $^1$University of California, Los Angeles,\ \ $^2$Google Research \\
    \texttt{telinwu@cs.ucla.edu}, \texttt{\{chgli,mingyang,taochen,spurthiah,bemike\}@google.com} \\
}
}

\newcommand\blfootnote[1]{%
  \begingroup
  \renewcommand\thefootnote{}\footnote{#1}%
  \addtocounter{footnote}{-1}%
  \endgroup
}

\begin{document}

\newcommand{\telin}[1]{{\color{blue}{\small\bf\sf [Te-Lin: #1]}}}
\newcommand{\all}[1]{{\color{orange}{\small\bf\sf [All: #1]}}}
\newcommand{\telinred}[1]{{\color{red}{\footnotesize [Te-Lin: #1]}}}

\newcommand{\cheng}[1]{{\color{red}{\small\bf\sf [Cheng: #1]}}}
\newcommand{\tao}[1]{{\color{cyan}{\small\bf\sf [Tao: #1]}}}
\newcommand{\mingyang}[1]{{\color{violet}{\small\bf\sf [Mingyang: #1]}}}
\newcommand{\spurthi}[1]{{\color{purple}{\small\bf\sf [Spurthi: #1]}}}
\newcommand{\mike}[1]{{\color{orange}{\small\bf\sf [Mike: #1]}}}

\newcommand{\SideNote}[2]{\todo[color=#1,size=\small]{#2}} 
\newcommand{\taoside}[1]{\SideNote{purple!40}{#1 --Tao}}
\newcommand{\mingyangside}[1]{\SideNote{blue!40}{#1 --Mingyang}}
\newcommand{\chengside}[1]{\SideNote{red!40}{#1 --Cheng}}
\newcommand{\spurthiside}[1]{\SideNote{green!40}{#1 --Spurthi}}
\newcommand{\mikeside}[1]{\SideNote{purple!40}{#1 --Mike}}
\newcommand{\telinside}[1]{\SideNote{pink!40}{#1 --Te-Lin}}

\newcommand{\Skip}[1]{}

\newcommand{\etal}{\textit{et al}.}
\newcommand{\ie}{\textit{i}.\textit{e}.\ }
\newcommand{\Ie}{\textit{I}.\textit{e}.\ }
\newcommand{\eg}{\textit{e}.\textit{g}.\ }
\newcommand{\Eg}{\textit{E}.\textit{g}.\ }
\newcommand{\lampret}{\textsc{LAMPreT}}
\newcommand{\lampretfull}{\textbf{L}ayout-\textbf{A}ware \textbf{M}ultimodal \textbf{PreT}raining}

\newcommand{\block}[1]{$\text{blk}_{#1}$}
\newcommand{\outrep}[1]{$\text{out}_{#1}$}
\newcommand{\blockrep}[1]{$\text{blkh}_{#1}$}
\newcommand{\clsi}[1]{$\text{CLS}_{#1}$}
\newcommand{\globalcls}{global-CLS}
\newcommand{\embd}[1]{$\text{embd}_{#1}$}

\newcommand{\tbref}[1]{Table \ref{#1}}
\newcommand{\equationref}[1]{Equation \ref{#1}}
\newcommand{\mypar}[1]{\noindent\textbf{#1}}

\newcommand{\tieconcat}{\mathbin{\mathpalette\dotieconcat\relax}}
\newcommand{\dotieconcat}[2]{
  \text{\raisebox{.8ex}{$\smallfrown$}}%
}

\maketitle

\textbf{\blfootnote{$^*$Work done during an internship at Google Research.}}

\begin{abstract}
    

Document layout comprises both structural and visual (\eg font-sizes) information that are vital but often ignored by machine learning models. The few existing models which do use layout information only consider  \textit{textual} contents, and overlook the existence of contents in other modalities such as images. Additionally, spatial interactions of presented contents in a layout was never fully exploited.

To bridge this gap, we parse a document into content blocks (\eg text, table, image) and propose a novel layout-aware multimodal hierarchical framework, LAMPreT, to model the blocks and the whole document. Our LAMPreT encodes each block with a multimodal transformer in the lower-level, and aggregates the block-level representations and connections utilizing a specifically designed transformer at the higher-level.
We design hierarchical pretraining objectives where the lower-level model is trained with the standard masked language modeling (MLM) loss and the image-text matching loss, and the higher-level model is trained with three layout-aware objectives:
(1) block-order predictions,
(2) masked block predictions, and
(3) image fitting predictions.
We evaluate the proposed model on two layout-aware tasks -- text block filling and image suggestion, and show the effectiveness of our proposed hierarchical architecture as well as pretraining techniques.

\end{abstract}

\section{Introduction}

Layout, the structural and visual presentation of the contents within a document, is a key aspect for writers to compose documents and for readers to understand documents. Specifically, the planning and the arrangements of how the contents are spatially structured, as well as the use of multiple modalities (\eg texts, graphics, tables), is highly influential in the choice of reading strategies from the readers. Hence, a well crafted layout can lead to better comprehension of the presented contents, and layout information is vital for document understanding~\cite{wright1999psychology,hartley2013designing}.





Learning a document-level representation with awareness of the layout
has started to draw attention in the research community, especially in achieving better semantic document understanding~\cite{katti2018chargrid, denk2019bertgrid, xu2020layoutlm}.
However, most prior works focus on rather surface forms of the layout, such as comprehending table hierarchy~\cite{wang2020docstruct} 
and the claimed multimodality being referred to OCR detected features of the textual components~\cite{hua2020attention, zhang2020trie}. Moreover, prior works mostly concern documents in the domain of scanned templated documents like receipts~\cite{pramanik2020towards}, not as \textit{content-rich} and \textit{layout-flexible} as articles such as Wikipedia pages.

In this paper, we propose~\lampretfull, dubbed~\lampret, aiming for a more general-purposed pretraining methodology which exploits both the structure and the content of documents, and considers multimedia contents, such as images, to learn a comprehensive multimodal document representation. 
Specifically, we utilize an in-house document tokenizer to parse HTML formatted pages into several \textit{content blocks}, where each \textit{block}
has the following features:
(1) spatial position,
(2) semantic types, \eg headers and tables, and
(3) attributes like font-sizes.

Inspired by the inherent hierarchy in the contents, our~\lampret~framework is hierarchical, consisting of two cascaded transformers~\cite{vaswani2017attention}. The lower-level transformer takes as inputs the parsed multimodal content blocks serialized by their sorted spatial positions, and the output \textit{block-level} representations are consumed by the higher-level transformer. The lower-level model is trained with the Masked Language Modeling (MLM) objective~\cite{devlin2019bert} and an image-to-text matching prediction for grounding different input modalities.
For training the higher-level model, we propose three novel \textit{block-level} pretraining objectives aiming to exploit the structure of a document:
(1) \textbf{block-ordering prediction} requires the model to predict whether the input blocks are properly ordered,
(2) \textbf{masked-block predictions} shares similar spirit with textual MLM but acts at the textual \textit{block-level}, and
(3) \textbf{image fitting predictions} requires the model to select the most suitable image for a missing image block.


We evaluate our proposed~\lampret~framework on two downstream document completion tasks:
(1) \textbf{Text block filling} which aims to select
the most appropriate textual block for a missing block to complete a document, 
and (2) \textbf{Image content suggestion} where the models are required to correctly retrieve the most suitable image at a layout position for a particular document, from a sizable set of candidates approximating realistic scenarios of composing documents.
We show the effectiveness of our~\lampret~framework and the benefits of incorporating multimodality, as well as conduct extensive ablation studies on its components.
Our main contributions are as follows:
\vspace{-.5em}
\begin{itemize}[leftmargin=*]
    \item To our best knowledge, we are the first to consider layout multimodality in the context of the interactions between the actual image and the textual contents within a document.
    \vspace{-.7em}
    \item Propose a hierarchical framework with structure-exploiting pretraining objectives to learn layout-aware document representations.
    \vspace{-.7em}
    \item Design novel downstream tasks to evaluate the layout-awareness of the learned document representations, with the hope to spur relevant future research in multimodal document understanding.
\end{itemize}

\section{Related Works}


\mypar{Document or Long-Text Learning.} 
The recent advancements in NLP with the help of transformers~\cite{vaswani2017attention}, has encouraged research in transformer-based models that can go beyond the previous maximally allowed input length in models such as BERT~\cite{devlin2019bert}.
Longformer~\cite{Beltagy2020Longformer}, Reformer~\cite{kitaev2020reformer}, and the recently proposed BigBird~\cite{zaheer2020big}, are all capable of handling much longer input texts even to the whole document-level.
The focus of this work, on one hand, is to design a framework that explicitly exploits the structure of the document contents, rather than modeling long document texts in the conventional way; on the other hand, we do not make any assumption of the base model used for~\lampret~and hence any of these recent models can replace our current base model, which is BERT.

\vspace{0.3em}

\mypar{Document Layouts.} 
Obtaining document layouts can be done by utilizing a conventional and rather rule-based technique, such as VIPS~\cite{cai2003vips}, and recent deep learning approaches~\cite{yang2017learning, soto2019visual, ling2020deeppapercomposer}, where the computer vision models~\cite{ren2015faster} is adopted.
CharGrid~\cite{katti2018chargrid} and its extensions~\cite{denk2019bertgrid, kerroumi2020visualwordgrid} assume the layout contents are visually interpreted via computer vision techniques such as OCR, and propose learning frameworks to semantically understand the documents from a 2D aspect.
Other prior works utilize document layouts as an effective component for information extractions~\cite{hua2020attention, gorai2020layout, yu2020pick}, and provide a benchmark for surface semantic understanding of documents~\cite{li2020docbank}.
Our work aims to go beyond surface understanding and exploit the layout explicitly in a modeling perspective in combined with the fine-grained language and vision models.

\vspace{0.3em}

\mypar{Multimodality.}
Multimodal grounding is an important paradigm for training visual-linguistics models.
In this work, we adapt the basic model configuration from recently proposed BERT-based visual-linguistics models~\cite{li2019visualbert, su2019vl, lu2019vilbert, chen2019uniter, li2020oscar, yu2020ernie} to fuse the textual and image modalities.
Specifically, we also adapt the image-to-text matching training objective inspired by these models as one of the components in~\lampret.
Prior works concern multimodality in learning or extracting document layouts~\cite{wang2020docstruct, pramanik2020towards, zhang2020trie, xu2020layoutlm} mostly by viewing a document as a structured imagery, while we model the actual multimodality in the contents such as how the texts should interact with the images within a document.


\section{Preliminaries}
\label{sec-prelims}

\begin{figure}[ht]
\centering
    \includegraphics[width=0.85\columnwidth]{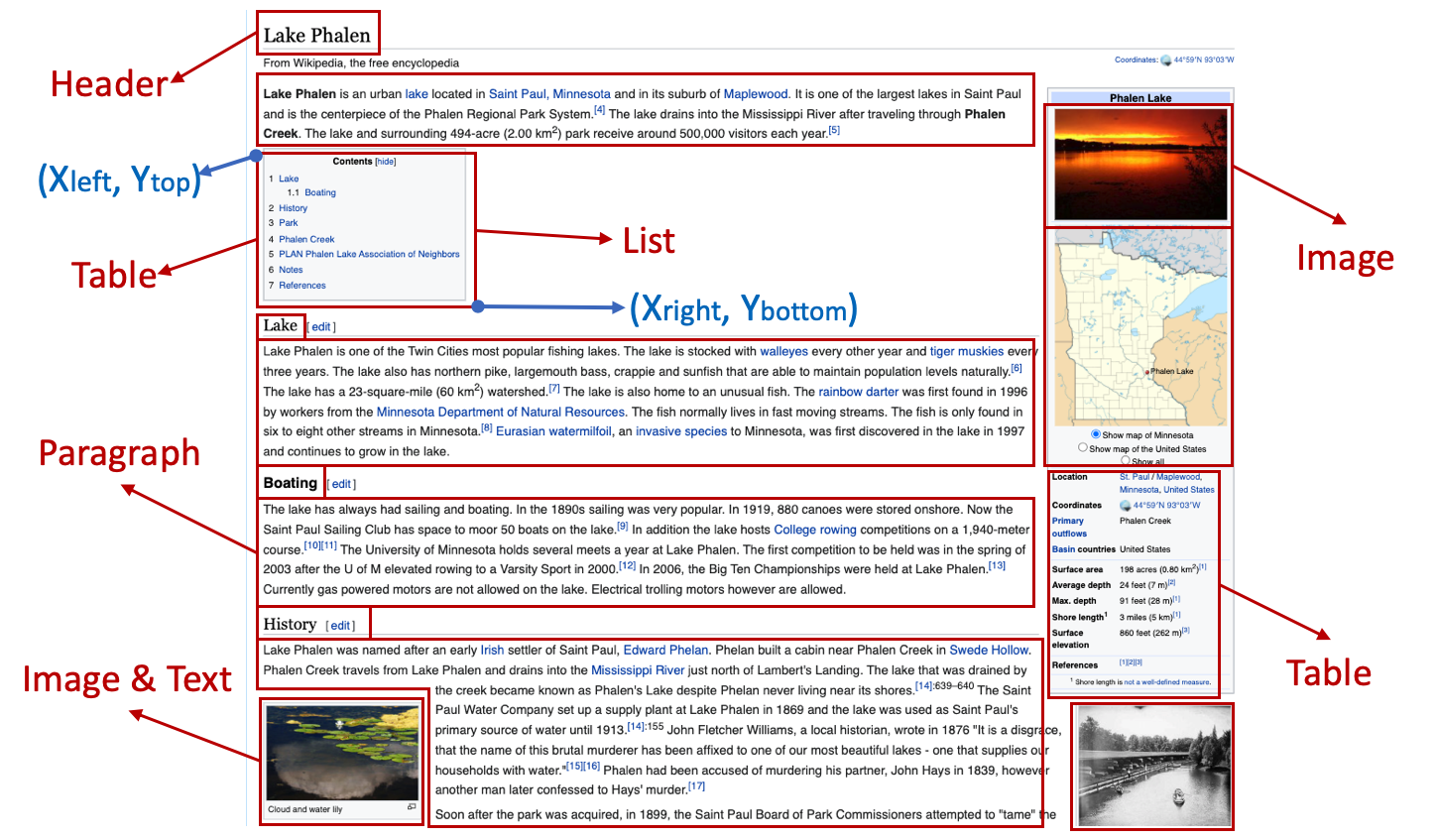}
    \caption{
        \footnotesize
       \textbf{An example page parsed by the document tokenizer.} Each \textcolor{red}{red box} indicates a content block. \textcolor{blue}{Blue colored coordinates} are an exemplar block position tuple (origin of the coordinate system is at \textit{top-left} corner of the entire document).
    }
    \label{fig:tokenizer}
\end{figure}


\mypar{Document Tokenizer:}
In this work, the layout is obtained by an in-house document parsing tool\footnote{Note that document parsing is not of our main focus, we assume our data is already parsed by the ready-to-use tool.}, which primarily handles HTML formatted web-page documents similarly to the aforementioned VIPS~\cite{cai2003vips}.
\Figref{fig:tokenizer} illustrates how a document is \textit{tokenized} (parsed) into several small \textit{content \textbf{blocks}}, each is a small proportion of the document which shows a clear spatial boundary to the others. Each block has the following features:
\begin{itemize}[leftmargin=*]
    \item \textbf{Block Position:} The 2D real valued position of the bounding box which encompasses the block, represented by XY coordinate tuples of (\textit{top-left}, \textit{bottom-right}) corners as illustrated in~\Figref{fig:tokenizer}. Each XY coordinate is normalized to $\in [0, 1]$.
    \item \textbf{Block Type:} The semantic type of the content presented in the block. There are in total 13 different types defined by our document tokenizer, with the most popular ones being header, paragraph, image, list (bullet-items), and table, etc.
    \item \textbf{Block Attributes:} The visual presentations of the \textbf{texts} featured in a block. Two types of attributes are generated by our tokenizer:
    (1) \textbf{scalar typed}, such as font-size, which is normalized to $\in [0, 1]$ with 1 indicating the largest possible font-size,
    and (2) \textbf{binary-typed}, such as indicating if the text is \textbf{bold}, \textit{italic}, or \underline{underlined}.
    \item \textbf{Multimedia:} There might be multimedia contents such as images, thumbnails, and videos in a block. In this work, we only consider images for our multimodal model. But our model can be easily extended to other multimedia contents.
    Note that multimedia content can be a block itself, as illustrated in~\Figref{fig:tokenizer}, the \textit{Image} block is different from the \textit{Image \& Text} block.
\end{itemize}

\begin{figure*}[ht!]
\begin{subtable}{.3\textwidth}
\centering
  \includegraphics[width=0.7\textwidth]{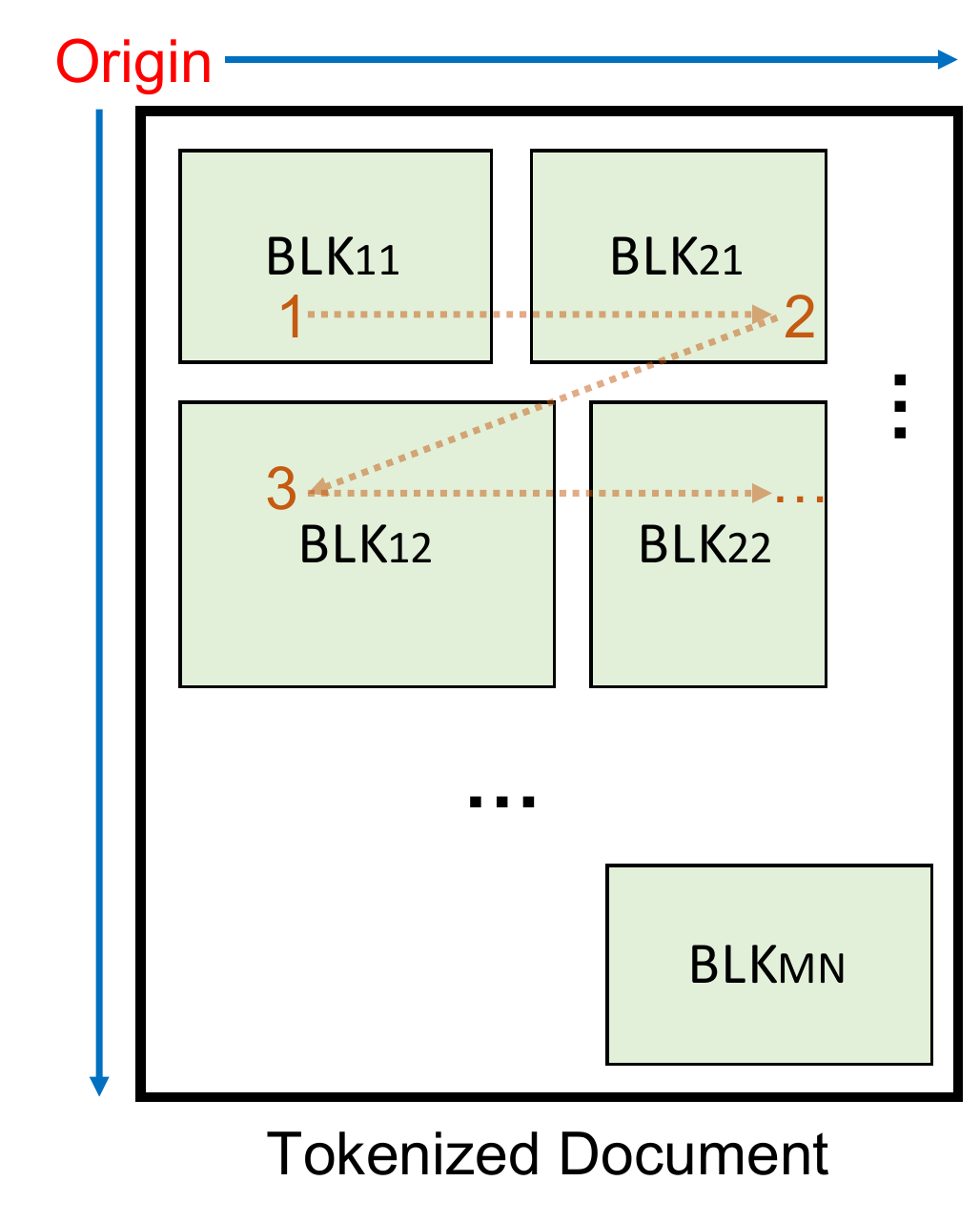}
    \caption{
        Layout Sorting.
    }
    \label{fig:2d_sort}
\end{subtable}%
\quad
\begin{subtable}{.7\textwidth}
\centering
  \includegraphics[width=0.9\textwidth]{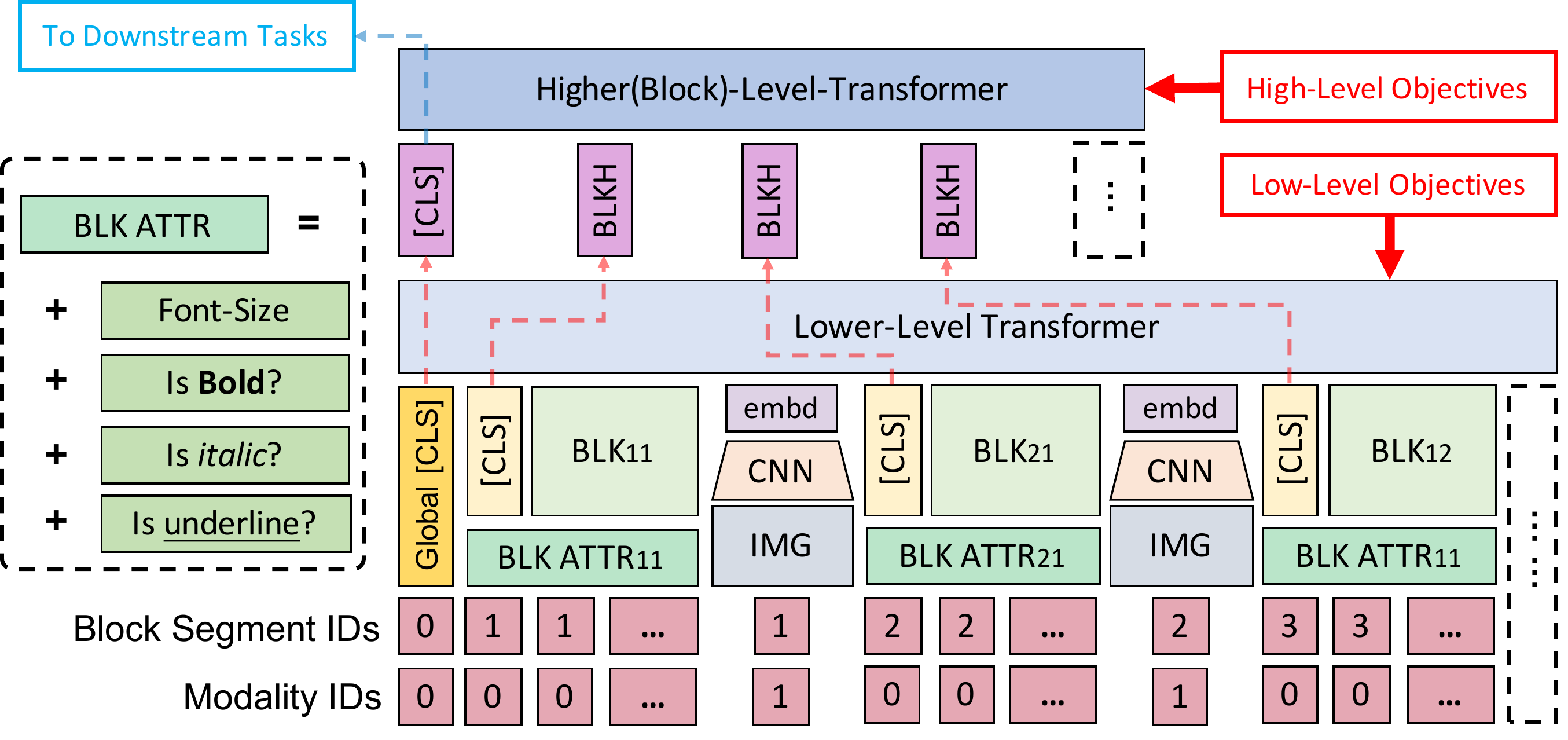}
    \caption{
        Hierarchical Architecture.
    }
    \label{fig:model_overview}
\end{subtable}
\caption{\footnotesize \textbf{~\lampret~Framework Overview:}
\textbf{(a) Sorting and serializing blocks:} Setting the \textit{origin} of the document at the top-left-most corner, we perform a 2D sorting on the content blocks anchoring around their \textit{top-left} coordinate. We then serialize the sorted blocks in a \textit{zigzag} fashion to obtain a reasonable ordering for the model inputs.
\textbf{(b) Hierarchical formulation:}~\lampret~framework exploits the inherent hierarchical nature of document layouts. The input representation of each block \block{i} contains the embeddings of Wordpiece tokens, block-segment-ids, modalities, and attributional features. The output representations of the lower-level model at each \clsi{i} position are fed to the higher-level model.
Different levels of objectives are applied to the models in different hierarchy, and the representations at \globalcls~are used in downstream tasks.
}
\label{tab:data_wsc}
\end{figure*}

\vspace{0.3em}
\mypar{Layout:}
We define layout as the structural presentation of the tokenized content blocks, \ie their relative positions and orders, and the aforementioned attributional features of the textual contents within a block.
In order to properly prepare the input representations for our models, we first \textit{\textbf{sort}} the tokenized content blocks, with respect to the two dimensional coordinates of their \textit{\textbf{top-left}} corners, as illustrated in~\Figref{fig:2d_sort}. We sort the Y-axis first and then X-axis, with the intuition that for the documents this work focuses on, the vertical order is slightly more important than the horizontal. The sorted blocks are serialized in a \textit{\textbf{zigzag}} fashion and then fed to the models.
Note that our proposed way of sorting is not necessarily the optimal one, where our main focus here is to provide a meaningful ordering for the models to take inputs.

\section{LAMPreT}


\newcolumntype{L}{>{\arraybackslash}m{5cm}}

\begin{table}[t!]
\centering
\small
    \scalebox{0.9}{
    \begin{tabular}{c|L}
    \toprule
    \textbf{Expression} & \multicolumn{1}{c}{\textbf{Descriptions}}\\
    \midrule
    \block{i} or \block{ij} & The $i$-th content block in the serialized order, or the $i$ and $j$-th in 2-dimension.  \\
    \cline{1-2} \\[-.8em]
    \outrep{i} & Final output representation of block $i$ after the higher-level model. \\
    \cline{1-2} \\[-.8em]
    \blockrep{i} & Block-level representation of block $i$ after the lower-level model, taken as input to the higher-level model. \\
    \cline{1-2} \\[-.8em]
    \clsi{i} & CLS token for the $i$-th block, mainly for separating the input blocks and aggregating the $i$-th block-level representation.\\
    \cline{1-2} \\[-.8em]
    \globalcls & The global CLS token which is prepended at the beginning of the inputs, where the representation obtained at this position is regarded as the overall representation of the whole document. \\
    \cline{1-2} \\[-.8em]
    \embd{feature} & The embedding for a particular \textit{feature}. \\
    \bottomrule
    \end{tabular}
    }
\caption{\footnotesize \textbf{Terminologies} used throughout the paper.}
\label{tab:term}
\end{table}

Our~\lampret~framework aims to:
(1) model the inherent hierarchical formulation of layout-structured documents, and
(2) exploit the structure alongside the actual document contents to learn the representation.
The frequently used terminologies throughout the paper are defined in~\tbref{tab:term}.

\subsection{Model Overview}
\label{sec-overview}

\mypar{Hierarchical Architecture:}
In order to better comprehend a document structured with a particular layout, we consider two level of layout hierarchical formulation. Specifically, the lower-level of the hierarchy refers to the contents of a block 
such as text and/or images, while the higher-level concerns how these blocks are spatially structured. We design a framework consisting of two cascaded transformers taking different levels of inputs of a given document, as illustrated in~\Figref{fig:model_overview}.
The lower-level model takes as inputs the raw parsed contents, where each \textit{content block} is placed at its \textbf{serialized sorted} position (represented by block-segment-id, recall~\Figref{fig:2d_sort}).
Each block contains the textual contents and potentially also a few images (can be more than one), making the lower-level model inherently multimodal. Each block \block{i} is prepended with a \clsi{i}\footnote{The CLS special token in~\citet{devlin2019bert} can be used for downstream tasks, we follow the similar formulation to prepend each block with its own CLS token so each of them can contribute to the training when required.} special token for indicating the boundary of block contents. We also prepend a \globalcls~token at the beginning of the inputs for obtaining \textit{document-level} representation. 
The higher-level model then takes as inputs the block-level representations \blockrep{i}, \ie the outputs of the lower-level model at each \clsi{i} position.

\begin{figure*}[ht!]
\centering
    \includegraphics[width=0.77\textwidth]{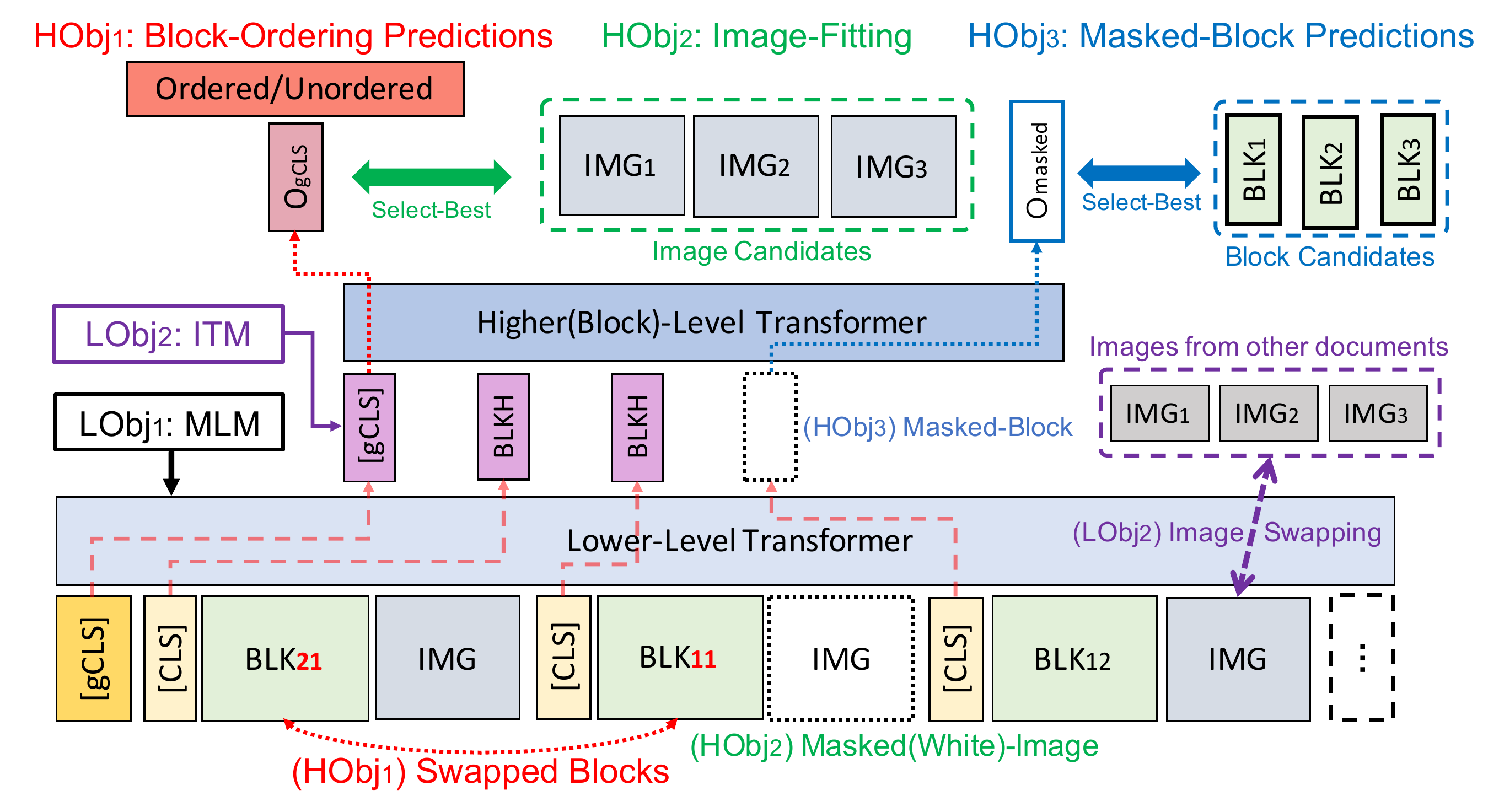}
    \caption{
        \footnotesize
        \textbf{LAMPreT Pretraining Objectives:}
        HObj$_{i}$ and LObj$_{i}$ denotes the $i$-th high- and low-level objective respectively. MLM and ITM stands for masked-language modeling and the image-text matching prediction for the low-level objectives.
        For each high-level objective, we illustrate: an exemplar block swapping for the block-ordering objective, an image masking for the image fitting objective, and a block masked at its \textit{block-level representation} for the block-MLM objective, respectively.
    }
    \label{fig:train_obj}
\end{figure*}

\vspace{0.3em}

\mypar{Input Representations:}
The textual contents are tokenized by the WordPiece~\cite{wu2016google} tokenizer. Each block is attached with a \textbf{block-segment-id} \textit{indexed by its serialized sorted position}, starting with 1 (0 is for the \globalcls~position). We map and round each real-valued font-size to an integer $\in [0, 10]$. The \textbf{boldness}, \underline{underline}, and \textit{italic} are simply represented as binary values $\in \{0, 1\}$. We also supplement a binary embedding indicating the modality.
The overall input representation for each token position is as follows:

\vspace{-1.em}

\begin{equation}
\begin{split}
\text{embd} = \text{embd}_{WordPiece} + \text{embd}_{block\_seg\_id} \\
              +\ \text{embd}_{type} + \text{embd}_{modality} + \text{embd}_{attr}
\end{split}
\end{equation}
where the \embd{attr} denotes the element-wise summed embedding from 
all the textual attributes.
For each block, we leave a design choice to truncate its contents with a maximally allowed token length, as well as a maximally allowed number of images.
More details are in the appendix~\Secref{a-input}.

\vspace{0.3em}

\mypar{Visual Embedding:}
The image contents are first fed to a convolutional neural network (CNN), followed by a transformation MLP (multi-layer perceptron) layer to align the resulting visual embedding \embd{img} to the same size of the textual token embedding.
For documents without any image contents, we simply pad the inputs with zero-image tensors, and the attention mask in the lower-level model is adjusted not to attend to those input positions.
For those standalone image blocks, we attach them to the closest text paragraph block (determined by the block positions) for a more straightforward block-level representation aggregation.

\subsection{Training Objectives}
\label{sec-objs}

\Figref{fig:train_obj} illustrates the training objectives on both the low- and the high-level of the~\lampret~framework. The lower-level training objectives aim to capture the finer-grained linguistics, visual information and the ability to handle multimodal inputs, while the higher-level training objectives aim to exploit the structural interactions among the contents at the \textit{block-level}.

\vspace{0.5em}

\mypar{Low-level objectives} for the lower-level model:
\begin{itemize}[leftmargin=*]
    
    \item \textbf{Masked Language Modeling (MLM):} Following BERT and its multimodal variants~\cite{lu2019vilbert,li2019visualbert}, we apply the MLM objective on one hand to further finetune the linguistics ability on our dataset, on the other hand to finetune language modeling with image modality. 
    
    \item \textbf{Image-Text Matching (ITM):} To further sharpen the model capability of handling multimodal inputs, we adapt the image-text matching (ITM) prediction used in~\cite{lu2019vilbert} to our setups. Specifically, for a given document $d$ containing one or more images, we sample a few candidate images from other documents $\{d'\}$ within the same mini-batch during training, and swap them with some images in $d$ with certain probabilities\footnote{With probability $\delta$ the images in $d$ are swapped, and $1-\delta$ the images in $d$ are remained the same.}. The model is then required to predict whether the textual contents match the resulting image sequences as binary classification.
    
\end{itemize}

\mypar{High-level objectives} for the higher-level model:
\begin{itemize}[leftmargin=*]

    \item \textbf{Block-Ordering Predictions (B-ORD):} Two input blocks are randomly selected and swapped\footnote{We limit the random re-ordering of the blocks to 2 to leave majority of other blocks untouched for other training objectives, and it is also empirically proven sufficiently effective.} (with certain probability remained) in their serialized order when inputting to the the lower-level model.
    An MLP\footnote{We include the specifications of all the MLPs used in this work in the appendix~\secref{a-mlps}.} which takes as input the output representation at the global CLS position, \outrep{\globalcls}, is trained to make the binary prediction on whether the input contents are following a proper order, \ie whether the two selected blocks are swapped. The block-segment-ids for the two selected blocks are replaced by a padding value to prevent the leak of the original order.
    
    \item \textbf{Block-MLM (B-MLM):} One or more textual blocks are masked out at their \textit{block-level} representations, \blockrep{i}, by replacing them with zero-tensors. The objective requires the model to \textit{select} the most suitable block for the masked position from a given set of candidate blocks, where the candidate set is constructed by collecting the blocks from all the documents (including self) within a mini-batch during training.
    An MLP layer then takes the concatenation of the output representations of the masked positions and the \textit{block-level} representations of the candidate blocks, \ie \textbf{concat}(\outrep{masked}, \blockrep{i}, \blockrep{j}, ...), and outputs the classification result of the index to the most suitable block. Since this objective is performed as a classification task, which requires a fixed number of candidates, in practice we truncate the candidates to a fixed number of blocks (with ground truth blocks deliberately included).
    
    \item \textbf{Image Fitting (IMG-FIT):} One or more images are masked out by replacing them with a \textit{mask-image-token}, which is a white image in our implementation. This objective requires the model to select the most suitable images from a set of candidate images for the masked-out images. Similar to Block-MLM, the candidate set is constructed by collecting the images from all the documents within a mini-batch during training, and a classification MLP layer is applied to predict the most suitable ones. The input to the MLP layer for the masked image in the $i$-th block is: \textbf{concat}(\outrep{\globalcls}, \outrep{\text{blk},i}, \embd{img,1}, \embd{img,2}, ...), where \embd{img, j} is the visual embedding of the $j$-th image candidate.
    We add the output representation at the \globalcls ~position to incorporate modeling the general trends of how the images are positioned within a document as it aggregates information from each block.
    In addition, a \textit{batch-level} mask is applied to filter out the losses of those data entries (documents) without any image contents.

\end{itemize}

\noindent ~\lampret~framework is jointly trained with a linear combination of the losses for the low- and high-level objectives:
\begin{equation}
\begin{split}
\mL_{\lampret} = \lambda_{1}L_{mlm} + \lambda_{2}L_{itm}\\
       +\ \lambda_{3}L_{b-ord} + \lambda_{4}L_{b-mlm} + \lambda_{5}L_{img-fit}
\end{split}
\end{equation}
where $\lambda_i$s are tunable hyperparameters\footnote{We set all $\lambda_i$s to 1 in our actual implementations.}, and all the losses $L$ are of classification cross-entropy loss.


\section{Experiments}


Our experiments aim to answer the following research questions: 
(1) Is the hierarchical formulation of~\lampret~effective?
(2) Are the proposed layout-aware training objectives effective, and how are they complementing one another?
(3) When and on what tasks does the multimodality help?

\subsection{Evaluation Tasks}


We design two \textit{document completion} tasks for evaluating the models:
(1) \textbf{text block filling} aims to evaluate the model capability of interpreting the structure of the textual contents, and
(2) \textbf{image suggestion} concerns the layout multimodality in addition to the structural aspect on texts.


\vspace{0.3em}

\mypar{Text Block Filling:} Suppose the 2D sorted and then serialized content blocks are $\{\text{blk}_1, \text{blk}_2, ..., \text{blk}_N\}$, we randomly select a block $\text{blk}_i$ to mask, and provide the context $\text{blk}_{1:i-1}$ as inputs to the model, while leaving $\text{blk}_{j:j+K} \cup \text{blk}_{i}$ as candidates, where $j > i$, $\text{blk}_{j}$ is spatially positioned after $\text{blk}_{i}$ by a certain margin (four rows in a common web-page document). The closest $K$ ($K$ = 5) blocks abide by the criteria are selected to make this task challenging.
The task is then to predict the correct block $\text{blk}_i$ from the candidate blocks. Note that this masked block $\text{blk}_i$ can be of any type of textual blocks, including header, an item from a list or table, etc. In this sense, the capability of prediction relies heavily on the understanding of the structure of the document.

\vspace{0.3em}

\mypar{Image Suggestion:} The model takes as inputs all the content blocks of a document with an image masked-out (by replacing it with all-white-image), and is required to predict the correct image from a given set of candidate images (including the ground truth of the masked out one). We extract $C$ ($C=1000$ in this work
) candidate images from documents unseen from the ones used during the pretraining for evaluating the models. Note that since this task aims to simulate suggesting the image contents when composing a novel document, for a more realistic setting, we \textbf{strip the textual blocks which encompass the direct captions to the images} in the dataset used for this task.

\vspace{0.3em}

\mypar{Finetuning on Downstream Tasks:}
To allow better fusion of low and high-level information, we have: $R_{doc} = \sigma(\alpha)\cdot \text{blkh}_{global-CLS} + (1-\sigma(\alpha))\cdot \text{out}_{global-CLS}$ to represent a document, where $\alpha$ is a task-dependent learnable scalar and $\sigma$ is the Sigmoid function.
We train an MLP to embed the document representation $R_{doc}$
to retrieve from a set of candidate embeddings $\{R_{cand}\}$ with a contrastive loss~\cite{hadsell2006dimensionality}, where $R_{cand, i} =$ \blockrep{i} for text block filling and $R_{cand, i} =$  \embd{img,i} for image suggestion.
Denote $Y$ as equals to $1$ if $R_{doc}$ is paired with the ground truth $R_{gt}$, and $0$ otherwise, and $m$ a predefined margin, we have:
\newcommand{\norm}[1]{\left\lVert#1\right\rVert}
\newcommand*{\Scale}[2][4]{\scalebox{#1}{$#2$}}%
\begingroup\makeatletter\def\f@size{9}\check@mathfonts
\begin{equation}
\begin{split}
D_{w}(R_{doc}, R_{cand, i}) = \norm{\textbf{\text{MLP}}(R_{doc}) - \textbf{\text{MLP}}(R_{cand, i})}_2 \\
L_{\text{contrastive}} = (1-Y)\frac{1}{2}(D_w)^2 + (Y)\frac{1}{2}\{max(0, m-D_w)\}^2
\end{split}
\end{equation}
\endgroup
\vspace{-1.em}

\vspace{-1em}

\subsection{Baselines}
\label{sec-baselines}

We compare our~\lampret~framework to the following baselines, more details are in~\Secref{a-baseline}: 

\vspace{0.5em}

\mypar{Single-Level LayoutLM:} The single-level, non-hierarchical variant of~\lampret~framework consists of only the lower-level model, which resembles the base model of the prior work LayoutLM~\cite{xu2020layoutlm}.
For training, both low-level objectives of~\lampret, MLM and ITM, are utilized. We compare against this baseline for examining the effectiveness of the hierarchical formulation and the high-level objectives.

\vspace{0.5em}

\mypar{CNN-Grid:} Inspired by prior work CharGrid and BERTGrid~\cite{katti2018chargrid, denk2019bertgrid}, we experiment replacing the transformer-based higher-level model with a CNN module (the original Char(BERT)Grid is applied on OCR detected results which does not fit our setup).
Each \textit{block-level} representation \blockrep{i} is inserted to a position on a 2D map according to the sorted 2D coordinates of the block. This results in a 3D tensor, where the original 1D \blockrep{i} representation becomes the channel dimension, and hence can be the input to a CNN module. While the output representation \outrep{\globalcls} of~\lampret~acts as the overall representation of the entire document, we apply an average pooling at the output of the CNN module to obtain the document-level representation of the same size. Since the CNN is viewed as a substitute of the higher-level model, we train this baseline with the same set of objectives of~\lampret, and hence the CNN-Grid baseline is the enhanced version of the prior works as we train it with additional layout-aware objectives.

\vspace{0.5em}

\mypar{Unimodality (Text-Only):} We are also interested in examining how our framework performs particularly for the text block filling task, if \textbf{the textual contents are the only} model inputs. We experiment the text-only variant of models for both our~\lampret~and the single-level LayoutLM.

\subsection{Implementation Details}

We initialize all the lower-level models (note that LayoutLM baseline is single-level)
with BERT-base-uncased
pretrained weights released from the original authors.
A pretrained CNN module is adopted for all the models and baselines to encode the images, and then transformed to the same embedding size of the Wordpiece token embedding, $768$, with an MLP layer.
For both block-MLM and image fitting pretraining objectives, we empirically select $20$ for the size of the candidate set, and hence the final output projection layers for these two objectives are $20$-way classification. 
For block-ordering, the output projection MLP is simply performing binary classification.
More implementation details are in the appendix~\Secref{a-impl}.

\subsection{Dataset}


In this work, we scrape a collection of English Wikipedia pages (Wikipages) for training and evaluating the models.
Our Wikipage dataset is uniformly sampled (scraped) from the entire Wikipedia, which makes it \textbf{diverse} and \textbf{rich} in the \textit{genres} and the \textit{contents}.
Furthermore, the Wikipage dataset naturally features \textbf{rich structures} as well as different \textbf{modalities} of contents.
We preprocess the collected Wikipages as described in~\secref{sec-prelims} to obtain content \textit{blocks}.
\tbref{tab:data-det} summarizes the essential dataset statistics.

\begin{table}[ht!]
\centering
\small
\scalebox{0.88}{
\begin{tabular}{lrr}
    \toprule
    \multicolumn{1}{c}{\textbf{Type}} & \multicolumn{2}{c}{\textbf{Counts}} \\
    \midrule
    \multicolumn{1}{c}{Total Pretraining Wikipages} & \multicolumn{2}{c}{6.2M} \\
    \multicolumn{1}{c}{Total Wikipages for Text Block Filling} & \multicolumn{2}{c}{30K} \\
    \multicolumn{1}{c}{Total Wikipages for Image Suggestion} & \multicolumn{2}{c}{28K} \\
    \multicolumn{1}{c}{Downstream Train / Val / Test} & \multicolumn{2}{c}{70\% / 10\% / 20\%} \\
    \toprule
    \multicolumn{1}{c}{\textbf{Type}} & \multicolumn{1}{c}{\textbf{Mean}} & \multicolumn{1}{c}{\textbf{Std}} \\
    \midrule
    \multicolumn{1}{l}{\# Blocks in a Wikipage} & 93.51 & 231.31 \\
    \multicolumn{1}{l}{\# Images in a Wikipage} & 0.46 & 1.33 \\
    \multicolumn{1}{l}{Document Token Length}  & 1083.02 & 1963.98 \\
    \bottomrule
\end{tabular}
}
\caption{\footnotesize \textbf{General statistics of the dataset.} 
Note that the Wikipages used in the downstream tasks are disjoint from the pretraining set.
For the image suggestion task, we only retain pages with at least one image.
}
\label{tab:data-det}
\end{table}

\begin{table*}[t]
\centering
\footnotesize
    \begin{tabular}{lcccccc}
    \toprule
    \multicolumn{1}{c}{\multirow{2}{*}{\textbf{Method}}} & \multirow{2}{*}{\textbf{Modality}} & \multicolumn{3}{c}{\textbf{Text Block Filling}} & \multicolumn{2}{c}{\textbf{Image Suggestion (C=1000)}}\\
    \cline{3-7} \\[-.8em]
    & & F1-Score & Precision (\%) & Recall (\%) & Recall @ 5 (\%) & MRR (\%)\\
    \midrule \\[-1.1em]
    
    CNN-Grid
    & Multimodal & 39.92 & 40.45 & 39.40  & 67.33 & 62.60 \\
    
    \midrule \\[-1.1em]
    
    \multirow{2}{*}{Single-Level LayoutLM}
    & Text-Only & 51.49 & 39.93 & \textbf{72.45} & --- & --- \\
    \cline{2-7} \\[-.8em]
    
    & Multimodal & 51.30 & 41.40 & 67.43  & 75.99 & 76.54 \\

    \midrule \\[-1.1em]

    \multirow{2}{*}{\lampret}
    & Text-Only & \textbf{52.36}* & \textbf{42.37}* & 68.50  & --- & --- \\
    \cline{2-7} \\[-.8em]
    
    & Multimodal & 52.09* & 41.85* & 68.98  & \textbf{99.98}* & \textbf{98.55}* \\
    
    \bottomrule
    \end{tabular}
\caption{\footnotesize \textbf{Model Performances:} Best performances for each metric is boldfaced. Our~\lampret~framework outperforms the carefully crafted baseline models in both tasks.
\textbf{* indicates that our~\lampret~framework is statistically significant compared with the best-performing baselines, according to the size of our test-set at level of 0.01.}}
\label{tab:model_perf}
\end{table*}

\begin{table*}[t]
\centering
\footnotesize
    \begin{tabular}{lccccc}
    \toprule
    \multicolumn{1}{c}{\multirow{2}{*}{\textbf{Ablated Components}}} & \multicolumn{3}{c}{\textbf{Text Block Filling}} & \multicolumn{2}{c}{\textbf{Image Suggestion}}\\
    \cline{2-6} \\[-.8em]
    & F1-Score & Precision (\%) & Recall (\%) & Recall @ 5 (\%) & MRR (\%)\\
    \midrule \\[-1.1em]
    
    w/o Image Fitting Objective
    & 50.72 & 38.82 & 39.40  & 90.63 & 89.68 \\
    
    w/o Block-MLM
    & 51.18 & 41.46 & 66.86  & 86.11 & 85.26 \\
    
    w/o block-ordering
    & 50.19 & 37.78 & 74.76  & 69.57 & 69.51 \\
    
    w/o Layout-Attributes
    & 50.98 & 40.39 & 69.07  & 96.42 & 95.27 \\
    
    \midrule \\[-1.1em]
    
    Multimodal~\lampret
    & 52.09 & 41.85 & 68.98  & 99.98 & 98.55 \\
    
    \bottomrule
    \end{tabular}
\caption{\footnotesize \textbf{Model Ablation Studies:} We examine the contribution of the high-level pretraining objectives on the multimodal version of~\lampret. Each row denotes the pretraining is conducted with the indicated objective excluded.
We also include an ablation on pretraining without the attributional features (last row), which is shown more effective on the text-based task.}
\label{tab:ablation}
\end{table*}

\vspace{-1.5em}

\vspace{1.2em}

\subsection{Evaluation Metrics}


Since the downstream tasks are trained with contrastive objective and performed in a retrieval fashion,
we adopt two common ranking-based metrics to quantify the model performances:

\vspace{0.3em}

\mypar{Mean Reciprocal Rank (MRR):} We compute the reciprocal (\ie the multiplicative inverse) ranks of the ground truth items in the given candidates list, and average them across the whole test set.

\vspace{0.3em}

\mypar{Recall @ K:} We compute the recall in the top-K ranked items by counting the number of the ground truth items in such a top-K candidate list. Since we only have one ground truth item for each example, the recall is binary existence divided by K.

\subsection{Experimental Results}

\tbref{tab:model_perf} summarizes the model performances on the two proposed downstream tasks.


\vspace{0.5em}

\mypar{Text Block Filling.}
For the text block filling downstream task, our proposed~\lampret~model outperforms the baselines in both the precision and F-1 score metrics. It is worth noting that for both~\lampret~and the single-level LayoutLM, the unimodal text-only version performs slightly better than the multimodal version. We hypothesize that such results can be attributed to the sub-optimal multimodal representation fusing, that it can be potentially alleviated with more sophisticated and finer-grained multimodal grounding paradigms. Among the models, CNN-Grid baselines performs the worst, of which the attention mechanism in transformers is hypothesized to capture the block-level interactions better.



\vspace{0.5em}

\mypar{Image Suggestion.}
We only conduct the image suggestion downstream task for multimodal models, as unimodal ones do not have access to image based features. Our~\lampret~achieves almost perfect performance for both metrics (99\%), while all the baseline models suffer significant performance degradation. The hierarchical formulation and the accompanying high-level objectives are proven effective in~\lampret~as compared to single-level LayoutLM.
The CNN-Grid baseline again generally performs the worst. 



\vspace{0.3em}

\mypar{Model Ablation Studies.}
We are interested in analyzing the contributions of the high-level pretraining objectives for different downstream tasks.
\tbref{tab:ablation} shows an ablation analysis on the multimodal version of our~\lampret~framework. At each row, we deduct (1) one of the high-level pretraining objectives, or (2) the layout attributional features. In general, the block-ordering objective is empirically proven quite effective for both downstream tasks, judged by the performance degradation when it is excluded during the pretraining. It is worth noting as well that the exclusion of the layout attributes do not cause that much deterioration compared to other pretraining objectives, which hypothetically implies that the exploitation of the structural information of layout designed in the proposed~\lampret~framework is relatively more effective.

Here we want to point out that we train for much more iterations during downstream finetuning for the image suggestion task (until performance convergence) when the model is not pretrained with the image fitting objective, otherwise with the same settings this downstream task does not work.


\section{Conclusions and Future Works}


We propose a multimodal layout-aware document representation learning framework, ~\lampret.
Once a document is parsed into several spatially structured \textit{content blocks}, we sort and serialize them in a 2D formulation. \lampret~aims to model the inherent hierarchical formulation of a document layout by two cascaded transformers.
The lower-level model is trained with MLM and ITM objectives, while the higher-level model is trained with three specifically designed layout-exploiting objectives.
We evaluate~\lampret~on two downstream tasks:
(1) text block filling, and
(2) image suggestion task.

For future works, we envision that (1) a more delicately designed training paradigm such as curriculum training for our hierarchical models can be beneficial, and (2) a more sophisticated multimodal grounding to the lower-level model can potentially alleviate the slightly worse performance in the text-based downstream tasks.
Furthermore, we hope to expand the target domains from Wikipages to other \textit{content-rich} documents such as news articles.



\clearpage

\bibliography{anthology,custom}
\bibliographystyle{acl_natbib}

\clearpage

\appendix

\section{More Details of The Layout}
\label{a-dataset}


We hereby include more details that could help understanding the concept of layout defined in this work and how to process it, as well as its essential building component, the \textit{content blocks}.

\subsection{Sorting The Blocks}

In our dataset, the block positions are structured in the form of a standard bounding box coordinate tuple:
$(X_{\text{left}}, Y_{\text{top}}, X_{\text{right}}, Y_{\text{bottom}})$, where the pair $(X_{\text{left}}, Y_{\text{top}}$) represents the coordinates of the \textit{top-left} corner, while the pair $(X_{\text{right}}, Y_{\text{bottom}})$ represents the \textit{bottom-right} corner of the bounding box of a particular content block. In this formulation, even non-rectangular bounding regions for contents can be properly represented and bounded, and our sorting paradigm does not overlook those blocks.

We sort the blocks anchoring around the \textit{top-left} corner of the box bounding the content block, in this way it is ensured their starting position is of the main consideration. Since even a semantically latter block can have earlier ending position if viewed from the document origin as illustrated in~\Figref{fig:2d_sort}, we refrain from using the \textit{bottom-right} corner for such a reason. We sort the two coordinates following the order: sort the $Y_{top}$ first and then $X_{left}$, with a standard sequence sorting algorithm. We then perform a \textit{zigzag} traversal to serialize the content blocks for the inputs to the models. Note that if the blocks are properly sorted (such as the aforementioned sorting), the serialization would reasonably preserve the relative orders in their original 2D formulations, and hence is a proper input sequence to a transformer-based model. The block-segment-ids defined after the serialization then acts similarly to the positional encoding of a transformer model, only that it is at the \textit{block-level}.

\subsection{Block Features}

\mypar{Block Attributes:}
For the real-value-typed textual attributional features, \ie font-size, our original data is prepared with a normalized value with the maximum indicating the H1 header size in a standard HTML web-page. We map this to a range of $\in [0, 10]$, and round the results to obtain integer values, so that we can regard these integer-scalars as font-size ids and use the standard embedding technique, \ie a linear layer (matrix) which retrieves the 1D embedding with the input ids.

\vspace{0.3em}

\mypar{Block Types:}
Each of the 13 block types is attached with an integer id which can then be used for standard embedding technique, starting from 0.
We use block type id $=13$, which is the $14$-th id for indicating padding block, which is used in the block-ordering prediction objective for preventing leaking order information, as mentioned in~\secref{sec-objs}.
We include~\tbref{tab:range} to summarize the ranges described above for all the discrete and/or discretized features of the inputs to our models.

\begin{table}[t]
\centering
\footnotesize
\begin{tabular}{lc}
    \toprule
     \multicolumn{1}{c}{\textbf{Feature}} & \textbf{Ranges or Bounds} \\
    \midrule
    Block Type & [0, 13] \\
    Font-Size & $\{0,1\}$, Binary \\
    Is Bold & $\{0,1\}$, Binary  \\
    Is Italic & $\{0,1\}$, Binary  \\
    Is Underline & $\{0,1\}$, Binary  \\
    Modality & $\{0,1\}$, Binary  \\
    Block-Segment-ID & [0, max. \# of blocks] \\
    WordPiece Vocabulary & [0, 30521] \\
    \bottomrule
\end{tabular}
\caption{\footnotesize \textbf{Ranges of discrete features:} For all the discrete and/or discretized features (\eg font-size), we include their ranges or bounds in this table. For the maximum number of blocks used in this work, please refer to~\tbref{tab:hyparams}.}
\label{tab:range}
\end{table}

\section{More Implementations Details}
\label{a-impl}

We implement all of the models (including all the baselines) in TensorFlow 1~\cite{abadi2016tensorflow}.
The BERT-base models which all the models based on, is adapted from the codes released in the original author code repository.
We also use the provided pretrained weights to initialize all of the models, specifically the lower-level models for those hierarchically formulated models.
The vocabulary size of the BERT-base model is 30522 according to the original configurations, where the input are tokenized by the WordPiece tokenizer.


\subsection{Input Representations}
\label{a-input}

In~\secref{sec-objs} in the main paper, we omit one of the obvious components used in the original BERT model, that is the token-level positional encoding, \ie \embd{positional}. Particularly for this positional embedding, we simply use a consecutive positional id scheme for all the inputs at their \textit{token-level}, starting from the first content block to the last and viewing the entire document contents as a single piece sentence segment. The image contents are also attached with a positional id without breaking the consecutive nature, we do not use different positional encoding schemes for different modalities and view them as the same piece of input for this encoding. 
As a result, the overall input representation is as follows:
\begin{equation}
\begin{split}
\text{embd} = \text{embd}_{WordPiece} + \text{embd}_{block\_seg\_id} \\
              +\ \text{embd}_{type} + \text{embd}_{modality} + \text{embd}_{attr} \\
              +\ \text{embd}_{positional}\ \ \ \ \ \ \ \ \ \ \ \ \ \ \ \ \ \ \ \ \ \ \ \ \ \ \ \ \ \ \ \ \ \ \ \
\end{split}
\end{equation}

\subsection{Higher-Level Model}
\label{a-higher}

For our~\lampret~framework, the architecture of our higher-level model is a $N$-layered transformer encoder, where the configuration of each layer of the encoder shares the same configurations used in BERT (where in BERT-base model, there are 12 this kind of layers cascaded sequentially). In our actual implementations, we use $N=3$ for our higher-level model.
All the other hyperparameters are remained the same as those in 1-layer BERT encoder (identical architecture for all 12 layers in the original BERT), where the hidden size is $768$ and the number of attention heads is $12$.

\subsection{Baseline Models}
\label{a-baseline}

\subsubsection{CNN-Grid}

\begin{figure*}[ht!]
\centering
    \includegraphics[width=0.8\textwidth]{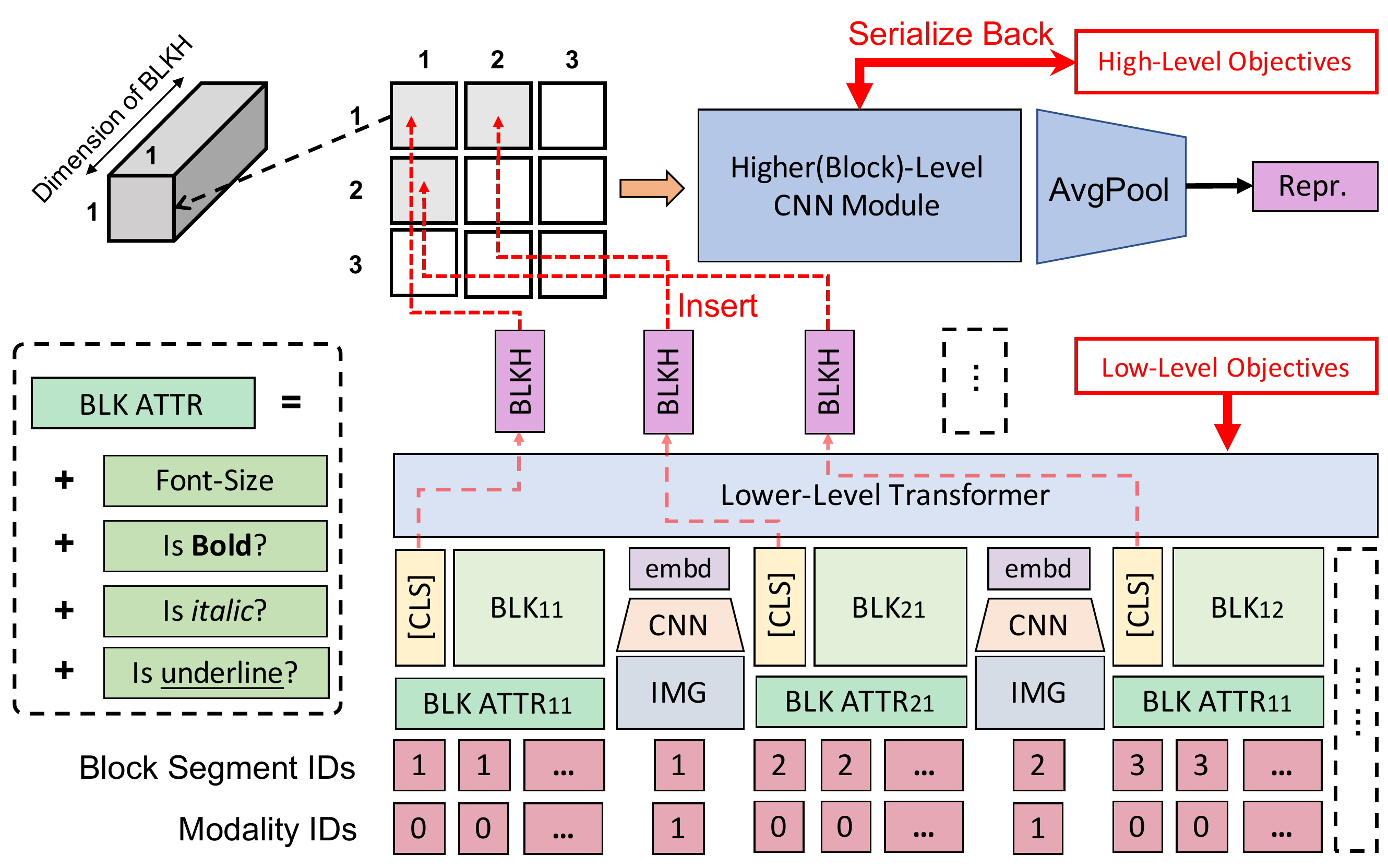}
    \caption{
        \footnotesize
        \textbf{Illustration of CNN-Grid baseline:} The block-level representations \blockrep{i} are inserted to the corresponding 2D positions of a 2D map, \eg \blockrep{21} of \block{21} is inserted to the position (2, 1) (X-axis is vertical and Y-axis is horizontal). This will result in a 3D tensor since as the dimension of~\blockrep{i} will be the \textit{channel dimension} of the inserted map. This 3D tensor is then fed to the convolutional modules, where the CNN module (represented as the rectangular shape) is designed to output the same shape of the input 3D tensor.
        Notice that the \globalcls~is not added since we do not need it for CNN higher-level model, instead, we perform an average pooling for reducing the height and width of the output tensor to $1\times1$ for obtaining a 1D \textit{document-level} representation.
        The high-level objectives can be applied to the combination of the \textit{document-level} representation and each of the~\outrep{i} after serializing the output tensor of the first stage CNN module back.
    }
    \label{fig:cnngrid}
\end{figure*}
\Figref{fig:cnngrid} illustrates the CNN-Grid baseline described in~\secref{sec-baselines} in the main paper. As can be seen, the lower-level model is identical to our proposed main framework~\lampret, while we use a convolutional architecture for the higher-level model instead of the (empirically proven better) transformer. For each \textit{block-level} representation~\blockrep{i}, we \textit{insert} it to a 2D map according to its sorted 2D XY-position, with the dimension of~\blockrep{i} being the third dimension of the 2D map, which eventually results in a 3D tensor. As a result, the constructed tensor should have the size of: (\text{Max. number of block}, \text{Max. number of block}, $\text{Dimension of blkh}_{i}$), \eg $(50, 50, 768)$ in this work. We then apply a $L$-layered CNN module to encode this 3D tensor, with kernel size = $3$ and $L=3$ to match the $N=3$ of the higher-level model of~\lampret. We adjust the kernel strides and paddings in the CNN module to ensure that the height and width of the encoded tensor remains same as the input 3D tensor, in this way we can serialize back the encoded 3D tensor following the same $zigzag$ fashion to perform the high-level pretraining objectives in~\lampret. However, since the CNN module is fundamentally different from the transformer model, we do not have a \globalcls~position, instead, we apply an average pooling to the outputs of the CNN module to reduce the height and width of the final encoded tensor to $1\times1$, a 1D representation as the higher-level document-level representation similar to~\outrep{\globalcls}. In combined with the aforementioned serialized back output representations, we can apply all the high-level objectives similar to our~\lampret~framework.

\subsubsection{Single-Level LayoutLM}

Although we call this baseline as \textit{LayoutLM}, it is still substantially different from the proposed model in~\cite{xu2020layoutlm}. The original LayoutLM utilizes an OCR-based vision parser on scanned documents, and hence the attributional visual presentations of the textual contents, such as the font-sizes and boldness, are implicitly embedded in the OCR-parsed contents. On the contrary, in our version of Single-Level LayoutLM, the block information and attributes are directly extracted from HTML elements, where the visual presentations of those attributes are explicitly encoded as several discrete variables. Another noticeable difference is that the original LayoutLM adopts 2D scalar-valued positional encoding from the four coordinates of the bounding boxes, while we first properly define a 2D sorting scheme and then serialize the content blocks in our layout so we only need to encode single-scalar block-segment-ids as the block-level positional encoding, with the fine-grained token-level positional encoding kept. Differing from the original LayoutLM, the MLM applied to our version of Single-Level LayoutLM is the standard linguistics-based masked learning, and we additionally incorporate the image-to-text matching prediction for aligning multimodal inputs, which is not concerned by the original LayoutLM work.

\subsection{MLP Layers}
\label{a-mlps}

\begin{table}[t]
\centering
\footnotesize
\scalebox{0.88}{
\begin{tabular}{lc}
    \toprule
     \multicolumn{1}{c}{\textbf{Description}} & \textbf{Output Dimension} \\
    \midrule
    Transformation layer for visual embedding & 768 \\
    MLP for ITM prediction & 2, binary \\
    MLP for Block-ordering & 2, binary \\
    MLP for Block-MLM & 20, classification \\
    MLP for Image-Fitting & 20, classification  \\
    MLP for downstream tasks & 64 \\
    \bottomrule
\end{tabular}
}
\caption{\footnotesize \textbf{Configurations of all the MLPs in this work:} The description indicates where the MLP layers are being used. We implement all of our MLP layers as one layered MLP and include the output dimension in this table.
}
\label{tab:mlps}
\end{table}

\tbref{tab:mlps} summarizes the configurations of all the MLP layers in this work. We list each MLP module mentioned in the main paper for better references on the implementation details.

\section{More Training Details}

\subsection{Hyperparameters}
\label{a-ssec:hyparams}

\begin{table*}[t]
\centering
\small
    \scalebox{0.8}{
    \begin{tabular}{lcccccccc}
    \toprule
    \multicolumn{1}{c}{\multirow{2}{*}{\textbf{Method}}} & \multirow{2}{*}{\textbf{Phase}} & \multirow{2}{*}{\textbf{Batch-Size}} & \multirow{2}{*}{\textbf{Initial LR}} & \textbf{\# Training} & \textbf{Max. Token} & \textbf{Max. Per-Block} & \textbf{Max.} & \multirow{2}{*}{\textbf{\# Params}} \\
    & & &  & \textbf{Iterations} & \textbf{Length} & \textbf{Token Length} & \textbf{\# blocks} &\\
    \midrule \\[-1.1em]
    
    \multirow{3}{*}{CNN-Grid}
    & Pretraining & 128 & $2 \times 10^{-5}$ & 600K & 512 & 50 & 50 & \multirow{3}{*}{120M} \\
    \cline{2-8} \\[-.8em]
    
    & Text Block Filling & 128 & $1 \times 10^{-6}$ & 30K  & 512 & 50 & 50 & \\
    & Image Suggestion & 128 & $1 \times 10^{-6}$ & 10K  & 512 & 50 & 50 & \\
    
    \midrule \\[-1.1em]
    
    \multirow{3}{*}{Single-Level LayoutLM}
    & Pretraining & 128 & $2 \times 10^{-5}$ & 600K & 512 & 50 & 50 & \multirow{3}{*}{110M} \\
    \cline{2-8} \\[-.8em]
    
    & Text Block Filling & 128 & $1 \times 10^{-6}$ & 30K  & 512 & 50 & 50 & \\
    & Image Suggestion & 128 & $1 \times 10^{-6}$ & 10K  & 512 & 50 & 50 & \\

    \midrule \\[-1.1em]

    \multirow{3}{*}{\lampret}
    & Pretraining & 128 & $2 \times 10^{-5}$ & 600K & 512 & 50 & 50 & \multirow{3}{*}{140M} \\
    \cline{2-8} \\[-.8em]
    
    & Text Block Filling & 128 & $1 \times 10^{-6}$ & 30K  & 512 & 50 & 50 & \\
    & Image Suggestion & 128 & $1 \times 10^{-6}$ & 10K  & 512 & 50 & 50 & \\
    
    \bottomrule
    \end{tabular}
    }
\caption{\footnotesize \textbf{Hyperparameters used for each model during different phases of training:} \textit{Initial LR} denotes initial learning rate. All the models are trained with Adam optimizers~\cite{kingma2014adam}. We include number of parameters of each model in the last column, denoted as \textit{\# params}. We use the same set of hyperparameters for the models involving different versions when using different modality of inputs. Particularly for the image suggestion downstream tasks, the models without the corresponding image-fitting pretraining objectives in the ablation studies in~\tbref{tab:ablation}, we train for 100K iterations, 10X the iterations used for those with the image-fitting pretraining objective, \ie 10K iterations as shown above. For the learning rate decay, we follow the procedure provided by the original BERT implementation.}
\label{tab:hyparams}
\end{table*}

\begin{table*}[t]
\centering
\footnotesize
\begin{tabular}{ccccc}
    \toprule
    \textbf{Type} & \textbf{Batch Size} & \textbf{Initial LR} & \textbf{\# Training Iterations} & \textbf{Max. \# blocks} \\
    \midrule
    \textbf{Bound (lower--upper)} & 32--256 & $5 \times 10^{-5}$--$1 \times 10^{-6}$ & 100K---600K & 20---50 \\
    \midrule
    \textbf{Number of Trials} & 6--10 & 2--3 & 2--4 & 8---10 \\
    \bottomrule
\end{tabular}
\caption{\footnotesize \textbf{Search bounds for hyperparameters:} for the hyperparameters of all the models.}
\label{tab:search}
\end{table*}

All the essential hyperparameters used throughout this work can be referred to in~\tbref{tab:hyparams}.
We also include the search bounds as well as the number of trials in searching for our manually-tuned hyperparameter search procedures in~\tbref{tab:search}.

\subsection{Hardware and Run-time}

We train all of our models (including all the baselines) on a set of TPU computing hardwares, similarly to the configuration for the original BERT model, \ie 4
Cloud TPUs in Pod configuration (16 TPU chips in total).
The run-time for the pretraining phase on average takes 2-3 days, and the run-time for the two downstream tasks take roughly
4 hours each.

\end{document}